\UseRawInputEncoding
\documentclass[10pt,twocolumn,letterpaper]{article}

\usepackage{cvpr}
\usepackage{times}
\usepackage{epsfig}
\usepackage{graphicx}
\usepackage{amsmath}
\usepackage{amssymb}

\usepackage{color}
\definecolor{note}{rgb}{0.1,0.1,1}

\usepackage{tikz,pgfplots}
\usepackage{flushend}

 \usepackage{colortbl}
\definecolor{rowblue}{rgb}{0.9,0.95,1}
\definecolor{rowgray}{rgb}{0.9275,0.9275,0.9275}
\definecolor{rowfgray}{rgb}{0.9275,0.9275,0.9275}
\definecolor{rowbgray}{rgb}{0.975,0.975,0.975}
\definecolor{rowdgray}{rgb}{0.85,0.85,0.85}

\usepackage[pagebackref=true,breaklinks=true,letterpaper=true,colorlinks,bookmarks=false]{hyperref}

\cvprfinalcopy 

\def\cvprPaperID{2793} 

\begin{document}
	

\title{BubbleNets: Learning to Select the Guidance Frame in \\Video Object Segmentation by Deep Sorting Frames} 

\author{Brent A. Griffin ~~~ Jason J. Corso \\
University of Michigan\\
{\tt\small \{griffb,jjcorso\}@umich.edu}
}

\maketitle

\begin{abstract}
	
	 Semi-supervised video object segmentation has made significant progress on real and challenging videos in recent years.
	 The current paradigm for segmentation methods and benchmark datasets is to segment objects in video provided a single annotation in the first frame.
	 However, we find that segmentation performance across the entire video varies dramatically when selecting an alternative frame for annotation.
	 This paper address the problem of learning to suggest the single best frame across the video for user annotation---this is, in fact, never the first frame of video.
	 We achieve this by introducing BubbleNets, a novel deep sorting network that learns to select frames using a  performance-based loss function that enables the conversion of expansive amounts of training examples from already existing datasets.
	 Using BubbleNets, we are able to achieve an 11\% relative improvement in segmentation performance on the DAVIS benchmark without any changes to the underlying method of segmentation.
   
\end{abstract}

\section{Introduction} 
	
Video object segmentation (VOS), the dense separation of objects in video from background, remains a hotly studied area of video understanding.
Motivated by the high cost of densely-annotated user segmentations in video \cite{DAVIS2018,ViGr09}, our community is developing many new VOS methods that are regularly evaluated on the benchmark datasets supporting VOS research \cite{SegTrackv2,DAVIS,DAVIS17,SegTrack,YTVOS}.
Compared to unsupervised VOS \cite{NLC,KEY,FST,WeSz17}, semi-supervised VOS, the problem of segmenting objects in video given a single user-annotated frame, has seen rampant advances, even within just the past year \cite{CINM,OSVOS,PML,FAVOS,SFL,VPN,CTN,OSVOS-S,RGMP,MSK,PLM,OSMN}.

\begin{figure}
	\centering
	\includegraphics[width=0.475\textwidth]{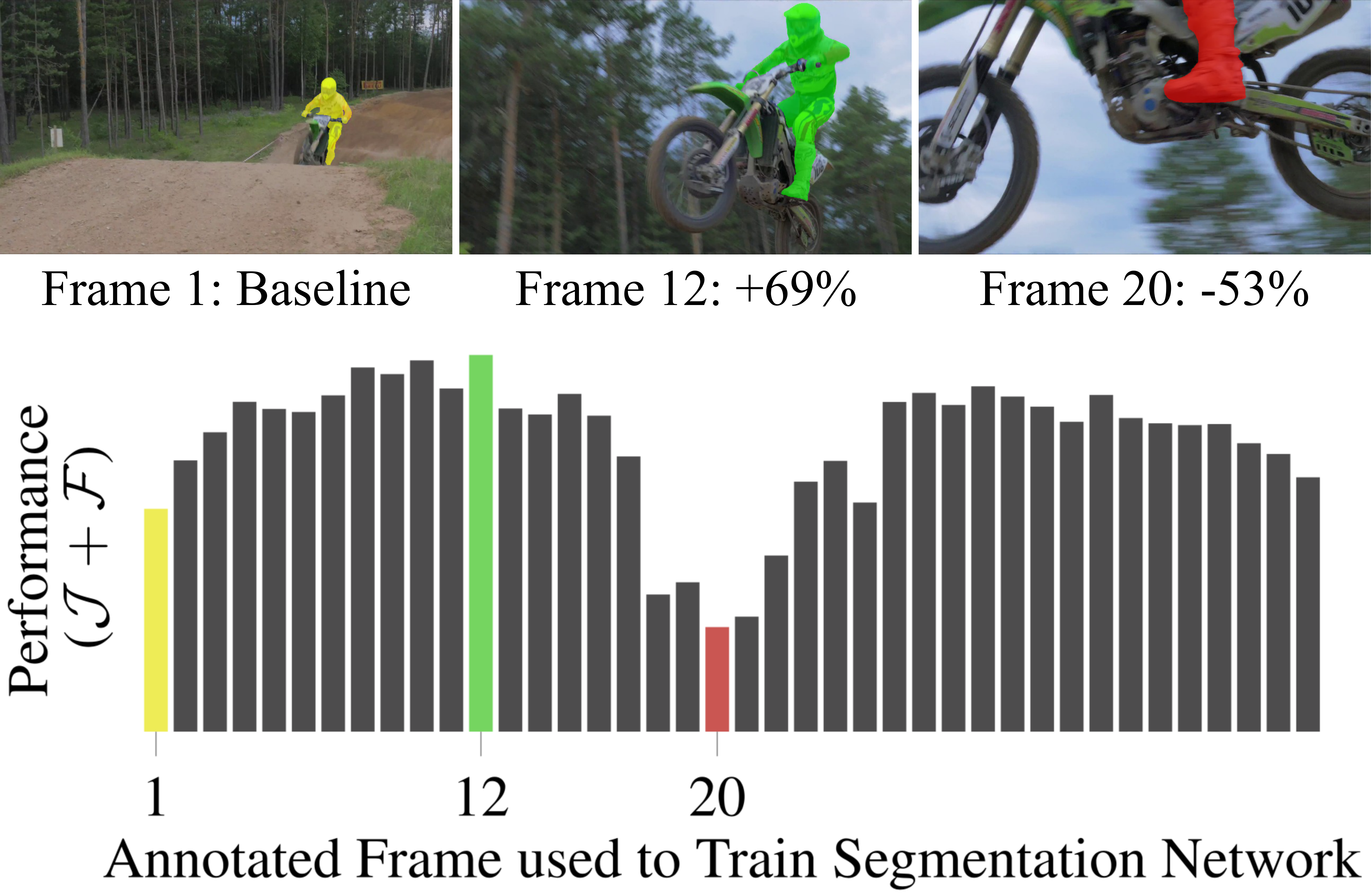}
	\caption{
		The current paradigm for video object segmentation is to segment an object annotated in the first frame of video (yellow, left). 
		However, selecting a different frame for annotation changes performance across the entire video [for better (green) \textit{or} worse (red)]. 
		To best use an annotator's time, our deep sorting framework suggests a frame that will improve segmentation performance.}
	\label{fig:frame_performance}
\end{figure}

The location and appearance of objects in video can change significantly from frame-to-frame, and, from our own analysis, we find that using different frames for annotation changes performance dramatically, as shown in Figure~\ref{fig:frame_performance}.
Annotating video data is an arduous process, so it is critical that we improve performance of semi-supervised VOS methods by providing the best single annotation frame possible.
However, we are not aware of any work that seeks to learn which frame to annotate for VOS.

To that end, this paper addresses the problem of selecting a single video frame for annotation that will lead to greater performance.
Starting from an untouched video, we select an annotation frame using our deep bubble sorting framework, which makes relative performance predictions between pairs of frames using our custom network, BubbleNets.
BubbleNets iteratively compares and swaps adjacent video frames until the frame with the greatest predicted performance is ranked highest, at which point, it is selected for the user to annotate and use for VOS.
To train BubbleNets, we use an innovative relative-performance-based loss that increases the number of training examples by orders of magnitude without increasing frame labeling requirements.
Finally, we evaluate BubbleNets annotation frame selection on multiple VOS datasets and achieve as much as an 11\% relative improvement in combined Jaccard measure and region contour accuracy ($\mathcal{J}+\mathcal{F}$) over the same segmentation method given first-frame annotations.

The first contribution of our paper is demonstrating the utility of alternative annotation frame selection strategies for VOS.
The current paradigm is to annotate an object in the first frame of video and then automatically segment that object in the remaining frames.
We provide thorough analysis across four datasets and identify simple frame-selection strategies that are immediately implementable for all VOS methods and lead to better performance than first-frame selection.
To the best of our knowledge, this represents the first critical investigation of segmentation performance for different annotation frame selection strategies.

The second contribution of our paper is the deep bubble sorting framework and corresponding implementation that improves VOS performance.
We are not aware of a single paper that investigates selection of the annotated frame in VOS.
The necessary innovation for our network-based approach is our loss formulation, which allows extensive training on relatively few initial examples.
We provide details on generating application-specific performance labels from pre-existing datasets, and our deep sorting formulation is general to all video processes that train on individual frames and have a measurable performance metric.
Using our custom network architecture and a modified loss function inspired by our VOS frame-selection analysis, we achieve the best frame-selection-based segmentation performance across all four evaluation datasets.

We provide source code for the current work at 
\url{https://github.com/griffbr/BubbleNets} and a video description at \url{https://youtu.be/0kNmm8SBnnU}.

\section{Related Work}

\subsection{Video Object Segmentation}

Multiple benchmarks are available to evaluate VOS methods, including: SegTrackv2 \cite{SegTrackv2,SegTrack}; DAVIS 2016, 2017, and 2018 \cite{DAVIS2018,DAVIS,DAVIS17}; and YouTube-VOS \cite{YTVOS}.
Moving away from the single-object hypothesis of DAVIS 2016, these datasets are increasingly focused on the segmentation of multiple objects, which increases the need for a user-provided annotation to specify each object of interest and has led to the development of more semi-supervised VOS methods using an annotated frame.
With some exceptions \cite{SEA,JMP,BVS,FCP}, the majority of semi-supervised VOS methods use an artificial neural network.

The amount of training data available for learning-based VOS methods has increased dramatically with the introduction of YouTube-VOS, which contains the most annotated frames of all current VOS benchmarks.
However,  due to the high cost of user annotation \cite{DAVIS2018,ViGr09}, YouTube-VOS only provides annotations for every fifth frame.
Operating on the assumption that every frame should be available to the user for annotation, we obtain training data from, and base the majority of our analysis from, DAVIS 2017, which contains the most training and validation examples of all fully annotated datasets and has many challenging video categories (e.g., occlusions, objects leaving view, appearance change, and multiple interacting objects).

For our BubbleNets implementation that selects annotated frames for VOS, we segment objects using One-Shot Video Object Segmentation (OSVOS) \cite{OSVOS},
which is state-of-the-art in VOS and has influenced other leading methods \cite{OSVOS-S,OnAVOS}.  
OSVOS uses a base network trained on ImageNet \cite{ImageNet} to recognize image features, re-trains on DAVIS 2016 to segment objects in video, and then fine-tunes the network for each video using a user-provided annotation.
One unique property of OSVOS is that it does not require temporal consistency, i.e., the order that OSVOS segments frames is inconsequential.
Conversely, even when segmentation methods operate sequentially \cite{CINM,VPN,DAVIS2017-2nd,DAVIS2017-1st,RGMP,MSK,OSMN}, segmentation can propagate forward \emph{and backward} from annotated frames selected later in a video.

\subsection{Active Learning}

Active learning (AL) is an area of research enabling learning algorithms to perform better with less training by letting them choose their own training data.
AL is especially useful in cases where large portions of data are unlabeled and manual labeling is expensive \cite{BeEtAl18}.
Selecting the best single annotated frame to train OSVOS represents a particularly hard problem in AL, starting to learn with no initial labeled instances, i.e., the cold start problem \cite{ColdStart98}.

Within AL, we are particularly interested in error reduction.
Error reduction is an intuitive sub-field that directly optimizes the objective of interest and produces more accurate learners with fewer labeled instances than uncertainty or hypothesis-based AL approaches \cite{Bu12}.
However, rather than going through all video frames and then formally predicting the expected error reduction associated with any one annotation frame, BubbleNets simplifies the problem by only comparing the relative performance of two frames at a time.
By combining our decision framework with a bubble sort, we iterate this selection process across the entire video and promote the frame with the best relative performance to be our selected  annotation frame.

Within computer vision, previous AL work includes measures to reduce costs associated with annotating images and selecting extra training frames \emph{after} using an initial set of user annotations.
Cost models predicting annotation times can be learned using a decision-theoretic approach \cite{ViGr09,ViJaGr10}.
Other work has focused on increasing the effectiveness of crowd-sourced annotations \cite{ViGr14}.
To improve tracking performance, active structured prediction has been used to suggest extra training frames after using an initial set of user annotations \cite{VoRa11}.
Within VOS, other work increases segmentation accuracy by having a user review segmentations and then add annotations on frames with poor performance \cite{OSVOS}. 
The DAVIS 2018 challenge includes emphasis on maximizing segmentation performance with decreased user annotation time \cite{DAVIS2018}.
In contrast, we are not estimating annotation costs or selecting extra annotation frames.
To support all semi-supervised VOS methods without increasing user effort, we are selecting a single frame for annotation that increases performance. 

\section{BubbleNets}

We design an artificial neural network, BubbleNets~(BN), that learns to suggest video frames for annotation that improve video object segmentation (VOS) performance.
To learn performance-based frame selection on our custom network, we generate our own labeled training data.
Labeled video data are expensive, so we design our network loss to learn from fewer initial frame labels, as discussed in Section~\ref{sec:BNrel}.
In Section~\ref{sec:deepsort}, we introduce our deep bubble sorting framework that uses BN performance predictions to select a single frame for annotation.
We provide details for our BN architecture in Section~\ref{sec:BNarch}.
In Section~\ref{sec:BNVOS}, we present our BN implementation for VOS with complete training and configuration details. 

\subsection{Predicting Relative Performance}
\label{sec:BNrel}

Assume we are given a set of $m$ training videos wherein each video has $n$ frames with labels corresponding to some performance metric, $y \in \mathbb{R}$, which we leave unspecified here but define in Section~\ref{sec:BNtrain}. 
Our goal is to learn to select the frame with the greatest performance from each video.

One way to accomplish this task is to use the entire video as input to a network (e.g., using an LSTM or 3D-ConvNet~\cite{CaZi17}) and output the frame index with the greatest predicted performance; however, this approach only has $m$ labeled training examples.
A second way to formulate this problem is to use individual frames as input to a network and output the predicted performance of each frame.
Using this formulation, the frame with the maximum predicted performance can be selected from each video and there are $m \times n$ labeled training examples.
While this is a significant improvement over $m$ examples, the second formulation only provides one training example per frame, which, for complicated and high annotation-cost processes like video object segmentation, makes the task of generating enough data to train a performance-prediction network impractical.

To that end, instead of directly estimating the predicted performance $y$ of each training frame, BN predicts the \emph{relative} difference in performance of two frames being compared (i.e., $y_i - y_j$ for frames $i$ and $j$ from the same video).
This difference may seem trivial, but it effectively increases the number of labels and training examples from $m \times n$ to $m \times {n \choose 2} \approx \frac{mn^2}{2}$. 

To further increase the number of unique training examples and increase BN's accuracy, we use $k$ random video reference frames as an additional network input.
When predicting the relative performance between two frames, additional consideration can be given to the frame that better represents the reference frames.
Thus, similar to architectures that process entire videos, reference frames provide some context for the video as a whole.
We find that reference frames not only increase BN's accuracy in practice but also increase the number of training examples from $m~\times~{n \choose 2}$ to $m \times {n \choose k + 2} \approx \frac{mn^{(k+2)}}{k+2}$.

Finally, we define our performance loss function as:
\begin{align}
\mathcal{L}(\textbf{W}) :=  (y_{i}-y_{j})-f(x_{i},x_{j},X_{\text{ref.}},\textbf{W}),
\label{eq:loss}
\end{align}
where $\textbf{W}$ are the trainable parameters of BN, $y_i$ is the performance label associated with the $i$th video frame, $x_i$ is the image and normalized frame index associated with the $i$th video frame,  $X_{\text{ref.}}$ is the set of $k$ reference images and frame indices,
and $f$ is the predicted relative performance.
For later use, denote the normalized frame index for the $i$th frame of an $n$-frame video as
\begin{align}
I_{i} = \frac{i}{n}.
\label{eq:idx}
\end{align}
Including $I$ as an input enables BN to also consider temporal proximity of frames for predicting performance.

\subsection{Deep Bubble Sorting}
\label{sec:deepsort}

Assume we train BubbleNets to predict the relative performance difference of two frames using the loss function \eqref{eq:loss} from Section~\ref{sec:BNrel}.
To select the frame with the greatest performance from a video, we use BN's relative performance predictions within a deep bubble sorting framework, iteratively comparing and swapping adjacent frames until we identify the frame with the greatest predicted relative (and overall) performance.

\begin{figure}[t!]
	\centering
	\includegraphics[width=0.475\textwidth]{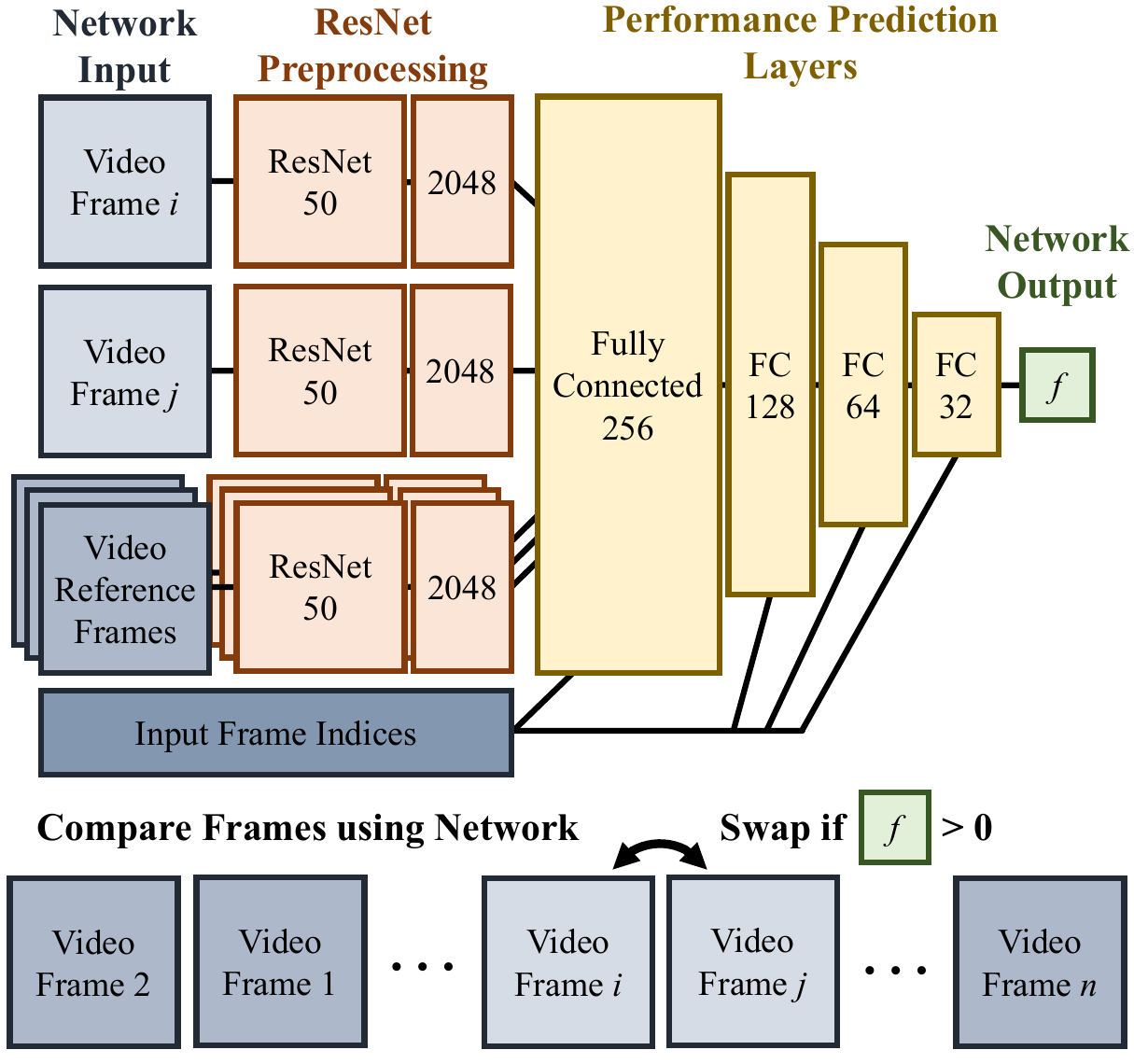}
	\caption{\textbf{BubbleNets Framework:} Deep sorting compares and swaps adjacent frames using their predicted relative performance.
	}
	\label{fig:BN}
\end{figure}

Our deep bubble sorting framework begins by comparing the first two video frames.
If BN predicts that the preceding frame has greater relative performance, the order of the two frames is swapped.
Next, the leading frame is compared (and potentially swapped) with the next adjacent frame, and this process passes forward until reaching the end of the video (see Figure~\ref{fig:BN}).
The frame ranked highest at the end of the sort is selected as the predicted best-performing frame. 

Normally, bubble sort is deterministic and only needs one pass through a list to promote the greatest element to the top; conversely, our deep bubble sorting framework is stochastic.
BN uses $k$ random video reference frames as input for each prediction, and using a different set of reference frames can change that prediction; thus, a BN comparison for the same two frames can change.
While bubble sort's redundancy is sub-optimal relative to other comparison sorts in many applications \cite{KN98}, revisiting previous comparisons is particularly effective given BN's stochastic nature.
Accordingly, our deep bubble sorting framework makes $n$ forward passes for an $n$-frame video, which is sufficient for a complete frame sort and increases the likelihood that the best-performing frame is promoted to the top.

One way to increase BN's consistency is to batch each network prediction over multiple sets of video reference frames.
By summing the predicted relative performance over the entire batch, we reduce the variability of each frame comparison.
However, two consequences of increasing batch size are: 1) increasing the chance of hitting a local minimum (i.e., some frame pairs are ordered incorrectly but never change) and 2) increasing execution time. 
In Section~\ref{sec:results}, we perform an ablation study to determine the best batch size for our specific application.

\begin{figure}[t!]
	\noindent\begin{minipage}{0.475\textwidth}
		\centering
\begin{tikzpicture}

\definecolor{color0}{rgb}{0.12156862745098,0.466666666666667,0.705882352941177}
\definecolor{color0}{rgb}{0.00000,0.306,0.596}
\definecolor{color3}{rgb}{0.3,0.85,0.3}
\definecolor{color0}{rgb}{1.00000,0.79610,0.01961} 
\definecolor{color0}{rgb}{0.00000,0.15290,0.29800} 

\definecolor{color0}{rgb}{0.3,0.3,0.3}
\definecolor{color0}{rgb}{0,0,0}

\begin{axis}[
width=8.5cm,
height=4cm,
tick align=outside,
tick pos=left,
x grid style={lightgray!92.02614379084967!black},
xmin=-1, xmax=43,
ymin=0, ymax=1.2,
ytick={0,0.2,0.4,0.6,0.8,1,1.2},
yticklabels={0.0,0.2,0.4,0.6,0.8,1.0,1.2},
ylabel={~~},
xlabel={Best Possible Sort},
xmajorticks=false,
xlabel near ticks,
axis line style={draw=none},
ymajorgrids,
grid style={lightgray!50},
ytick=\empty,
yticklabels={},
ymin=0, ymax=1.2,
/tikz/inner sep to outer sep/.style={inner sep=0pt, outer sep=.3333em},
x tick label style=inner sep to outer sep,
x label style=inner sep to outer sep,
y label style=inner sep to outer sep,
]
\draw[fill=color0,draw opacity=0] (axis cs:-0.4,0) rectangle (axis cs:0.4,0.415646325744935);
\draw[fill=color0,draw opacity=0] (axis cs:0.6,0) rectangle (axis cs:1.4,0.41661486698677);
\draw[fill=color0,draw opacity=0] (axis cs:1.6,0) rectangle (axis cs:2.4,0.428758490413158);
\draw[fill=color0,draw opacity=0] (axis cs:2.6,0) rectangle (axis cs:3.4,0.431305657422814);
\draw[fill=color0,draw opacity=0] (axis cs:3.6,0) rectangle (axis cs:4.4,0.440790943049437);
\draw[fill=color0,draw opacity=0] (axis cs:4.6,0) rectangle (axis cs:5.4,0.449374279715174);
\draw[fill=color0,draw opacity=0] (axis cs:5.6,0) rectangle (axis cs:6.4,0.460917790712668);
\draw[fill=color0,draw opacity=0] (axis cs:6.6,0) rectangle (axis cs:7.4,0.467868712899152);
\draw[fill=color0,draw opacity=0] (axis cs:7.6,0) rectangle (axis cs:8.4,0.480085896269366);
\draw[fill=color0,draw opacity=0] (axis cs:8.6,0) rectangle (axis cs:9.4,0.515851489303949);
\draw[fill=color0,draw opacity=0] (axis cs:9.6,0) rectangle (axis cs:10.4,0.537531091318991);
\draw[fill=color0,draw opacity=0] (axis cs:10.6,0) rectangle (axis cs:11.4,0.583032669744609);
\draw[fill=color0,draw opacity=0] (axis cs:11.6,0) rectangle (axis cs:12.4,0.585848168647253);
\draw[fill=color0,draw opacity=0] (axis cs:12.6,0) rectangle (axis cs:13.4,0.592496958541798);
\draw[fill=color0,draw opacity=0] (axis cs:13.6,0) rectangle (axis cs:14.4,0.666939038799613);
\draw[fill=color0,draw opacity=0] (axis cs:14.6,0) rectangle (axis cs:15.4,0.671824182147869);
\draw[fill=color0,draw opacity=0] (axis cs:15.6,0) rectangle (axis cs:16.4,0.697402865826715);
\draw[fill=color0,draw opacity=0] (axis cs:16.6,0) rectangle (axis cs:17.4,0.732774806607501);
\draw[fill=color0,draw opacity=0] (axis cs:17.6,0) rectangle (axis cs:18.4,0.770360829396125);
\draw[fill=color0,draw opacity=0] (axis cs:18.6,0) rectangle (axis cs:19.4,0.794219635420179);
\draw[fill=color0,draw opacity=0] (axis cs:19.6,0) rectangle (axis cs:20.4,0.829706692423229);
\draw[fill=color0,draw opacity=0] (axis cs:20.6,0) rectangle (axis cs:21.4,0.833482481157829);
\draw[fill=color0,draw opacity=0] (axis cs:21.6,0) rectangle (axis cs:22.4,0.847584694279896);
\draw[fill=color0,draw opacity=0] (axis cs:22.6,0) rectangle (axis cs:23.4,0.874774641140906);
\draw[fill=color0,draw opacity=0] (axis cs:23.6,0) rectangle (axis cs:24.4,0.911589362286067);
\draw[fill=color0,draw opacity=0] (axis cs:24.6,0) rectangle (axis cs:25.4,0.916030567453168);
\draw[fill=color0,draw opacity=0] (axis cs:25.6,0) rectangle (axis cs:26.4,0.94460697847888);
\draw[fill=color0,draw opacity=0] (axis cs:26.6,0) rectangle (axis cs:27.4,0.963116367523077);
\draw[fill=color0,draw opacity=0] (axis cs:27.6,0) rectangle (axis cs:28.4,0.969801641516352);
\draw[fill=color0,draw opacity=0] (axis cs:28.6,0) rectangle (axis cs:29.4,0.974727360979655);
\draw[fill=color0,draw opacity=0] (axis cs:29.6,0) rectangle (axis cs:30.4,0.977525214406065);
\draw[fill=color0,draw opacity=0] (axis cs:30.6,0) rectangle (axis cs:31.4,0.987967705091992);
\draw[fill=color0,draw opacity=0] (axis cs:31.6,0) rectangle (axis cs:32.4,1.00148154220955);
\draw[fill=color0,draw opacity=0] (axis cs:32.6,0) rectangle (axis cs:33.4,1.01588559560419);
\draw[fill=color0,draw opacity=0] (axis cs:33.6,0) rectangle (axis cs:34.4,1.02686978292214);
\draw[fill=color0,draw opacity=0] (axis cs:34.6,0) rectangle (axis cs:35.4,1.03789743221177);
\draw[fill=color0,draw opacity=0] (axis cs:35.6,0) rectangle (axis cs:36.4,1.04323694633995);
\draw[fill=color0,draw opacity=0] (axis cs:36.6,0) rectangle (axis cs:37.4,1.05851743899296);
\draw[fill=color0,draw opacity=0] (axis cs:37.6,0) rectangle (axis cs:38.4,1.07371483415905);
\draw[fill=color0,draw opacity=0] (axis cs:38.6,0) rectangle (axis cs:39.4,1.07639336582643);
\draw[fill=color0,draw opacity=0] (axis cs:39.6,0) rectangle (axis cs:40.4,1.09659098614273);
\draw[fill=color0,draw opacity=0] (axis cs:40.6,0) rectangle (axis cs:41.4,1.11993041844218);
\draw[fill=color3,draw opacity=0] (axis cs:41.6,0) rectangle (axis cs:42.4,1.12841265142064);

\end{axis}

\end{tikzpicture}
	\end{minipage}%
	\hfill
	\noindent\begin{minipage}{0.475\textwidth}
		\centering
\begin{tikzpicture}

\definecolor{color0}{rgb}{0.00000,0.15290,0.29800}
\definecolor{color1}{rgb}{0.85,0.3,0.3}
\definecolor{color2}{rgb}{0.93,0.93,0.15}
\definecolor{color3}{rgb}{0.3,0.85,0.3}
\definecolor{color4}{rgb}{0.3,0.3,0.85}

\definecolor{color0}{rgb}{1.00000,0.79610,0.01961} 

\begin{axis}[
width=8.5cm,
height=4cm,
tick align=outside,
tick pos=left,
x grid style={lightgray!92.02614379084967!black},
xmin=-1, xmax=43,
ymin=0, ymax=1.2,
ytick={0,0.2,0.4,0.6,0.8,1,1.2},
yticklabels={0.0,0.2,0.4,0.6,0.8,1.0,1.2},
ylabel={~~},
xlabel={BubbleNets Sort},
xmajorticks=false,
xlabel near ticks,
axis line style={draw=none},
ymajorgrids,
grid style={lightgray!50},
ytick=\empty,
yticklabels={},
ymin=0, ymax=1.2,
/tikz/inner sep to outer sep/.style={inner sep=0pt, outer sep=.3333em},
x tick label style=inner sep to outer sep,
x label style=inner sep to outer sep,
y label style=inner sep to outer sep,
]
\draw[fill=color0,draw opacity=0] (axis cs:-0.4,0) rectangle (axis cs:0.4,0.428758490413158);
\draw[fill=color0,draw opacity=0] (axis cs:0.6,0) rectangle (axis cs:1.4,0.41661486698677);
\draw[fill=color0,draw opacity=0] (axis cs:1.6,0) rectangle (axis cs:2.4,0.415646325744935);
\draw[fill=color0,draw opacity=0] (axis cs:2.6,0) rectangle (axis cs:3.4,0.431305657422814);
\draw[fill=color0,draw opacity=0] (axis cs:3.6,0) rectangle (axis cs:4.4,0.440790943049437);
\draw[fill=color0,draw opacity=0] (axis cs:4.6,0) rectangle (axis cs:5.4,0.460917790712668);
\draw[fill=color0,draw opacity=0] (axis cs:5.6,0) rectangle (axis cs:6.4,0.467868712899152);
\draw[fill=color0,draw opacity=0] (axis cs:6.6,0) rectangle (axis cs:7.4,0.480085896269366);
\draw[fill=color0,draw opacity=0] (axis cs:7.6,0) rectangle (axis cs:8.4,0.515851489303949);
\draw[fill=color0,draw opacity=0] (axis cs:8.6,0) rectangle (axis cs:9.4,0.537531091318991);
\draw[fill=color0,draw opacity=0] (axis cs:9.6,0) rectangle (axis cs:10.4,0.449374279715174);
\draw[fill=color0,draw opacity=0] (axis cs:10.6,0) rectangle (axis cs:11.4,0.592496958541798);
\draw[fill=color0,draw opacity=0] (axis cs:11.6,0) rectangle (axis cs:12.4,0.585848168647253);
\draw[fill=color0,draw opacity=0] (axis cs:12.6,0) rectangle (axis cs:13.4,0.671824182147869);
\draw[fill=color0,draw opacity=0] (axis cs:13.6,0) rectangle (axis cs:14.4,0.583032669744609);
\draw[fill=color0,draw opacity=0] (axis cs:14.6,0) rectangle (axis cs:15.4,0.697402865826715);
\draw[fill=color0,draw opacity=0] (axis cs:15.6,0) rectangle (axis cs:16.4,0.770360829396125);
\draw[fill=color0,draw opacity=0] (axis cs:16.6,0) rectangle (axis cs:17.4,0.732774806607501);
\draw[fill=color0,draw opacity=0] (axis cs:17.6,0) rectangle (axis cs:18.4,0.666939038799613);
\draw[fill=color0,draw opacity=0] (axis cs:18.6,0) rectangle (axis cs:19.4,0.829706692423229);
\draw[fill=color0,draw opacity=0] (axis cs:19.6,0) rectangle (axis cs:20.4,0.794219635420179);
\draw[fill=color0,draw opacity=0] (axis cs:20.6,0) rectangle (axis cs:21.4,0.916030567453168);
\draw[fill=color0,draw opacity=0] (axis cs:21.6,0) rectangle (axis cs:22.4,0.847584694279896);
\draw[fill=color0,draw opacity=0] (axis cs:22.6,0) rectangle (axis cs:23.4,0.833482481157829);
\draw[fill=color0,draw opacity=0] (axis cs:23.6,0) rectangle (axis cs:24.4,0.977525214406065);
\draw[fill=color0,draw opacity=0] (axis cs:24.6,0) rectangle (axis cs:25.4,1.00148154220955);
\draw[fill=color0,draw opacity=0] (axis cs:25.6,0) rectangle (axis cs:26.4,0.974727360979655);
\draw[fill=color0,draw opacity=0] (axis cs:26.6,0) rectangle (axis cs:27.4,0.874774641140906);
\draw[fill=color0,draw opacity=0] (axis cs:27.6,0) rectangle (axis cs:28.4,0.911589362286067);
\draw[fill=color0,draw opacity=0] (axis cs:28.6,0) rectangle (axis cs:29.4,0.963116367523077);
\draw[fill=color0,draw opacity=0] (axis cs:29.6,0) rectangle (axis cs:30.4,1.04323694633995);
\draw[fill=color0,draw opacity=0] (axis cs:30.6,0) rectangle (axis cs:31.4,1.02686978292214);
\draw[fill=color0,draw opacity=0] (axis cs:31.6,0) rectangle (axis cs:32.4,0.969801641516352);
\draw[fill=color0,draw opacity=0] (axis cs:32.6,0) rectangle (axis cs:33.4,1.01588559560419);
\draw[fill=color0,draw opacity=0] (axis cs:33.6,0) rectangle (axis cs:34.4,0.94460697847888);
\draw[fill=color0,draw opacity=0] (axis cs:34.6,0) rectangle (axis cs:35.4,1.07639336582643);
\draw[fill=color0,draw opacity=0] (axis cs:35.6,0) rectangle (axis cs:36.4,1.07371483415905);
\draw[fill=color0,draw opacity=0] (axis cs:36.6,0) rectangle (axis cs:37.4,1.09659098614273);
\draw[fill=color0,draw opacity=0] (axis cs:37.6,0) rectangle (axis cs:38.4,1.05851743899296);
\draw[fill=color0,draw opacity=0] (axis cs:38.6,0) rectangle (axis cs:39.4,0.987967705091992);
\draw[fill=color0,draw opacity=0] (axis cs:39.6,0) rectangle (axis cs:40.4,1.03789743221177);
\draw[fill=color0,draw opacity=0] (axis cs:40.6,0) rectangle (axis cs:41.4,1.11993041844218);
\draw[fill=color3,draw opacity=0] (axis cs:41.6,0) rectangle (axis cs:42.4,1.12841265142064);

\end{axis}

\end{tikzpicture}
	\end{minipage}%
	\hfill
	\noindent\begin{minipage}{0.475\textwidth}
		\centering
\begin{tikzpicture}

\definecolor{color0}{rgb}{0.3,0.3,0.3}
\definecolor{color1}{rgb}{0.7,0.15,0.15}
\definecolor{color2}{rgb}{0.93,0.93,0.15}
\definecolor{color3}{rgb}{0.3,0.85,0.3}
\definecolor{color4}{rgb}{0.3,0.3,0.85}

\definecolor{color0}{rgb}{0.00000,0.15290,0.29800} 
\definecolor{color0}{rgb}{0.3,0.3,0.3}

\begin{axis}[
width=8.5cm,
height=4cm,
tick align=outside,
tick pos=left,
x grid style={lightgray!92.02614379084967!black},
xmin=-1, xmax=43,
ymin=0, ymax=1.2,
ytick={0,0.2,0.4,0.6,0.8,1,1.2},
yticklabels={0.0,0.2,0.4,0.6,0.8,1.0,1.2},
ylabel={$\mathcal{J} + \mathcal{F}$},
ylabel near ticks,
xlabel={Initial Video Frame Indices},
xlabel near ticks,
xtick={0,9,19,29,39},
xticklabels={1,10,20,30,40},
axis line style={draw=none},
ymajorgrids,
grid style={lightgray!50},
ymin=0, ymax=1.2,
/tikz/inner sep to outer sep/.style={inner sep=0pt, outer sep=.3333em},
y label style=inner sep to outer sep,
x tick label style=inner sep to outer sep,
x label style={inner sep=7pt, outer sep=.3333em},
]
\draw[fill=color0,draw opacity=0] (axis cs:-0.4,0) rectangle (axis cs:0.4,0.428758490413158);
\draw[fill=color0,draw opacity=0] (axis cs:0.6,0) rectangle (axis cs:1.4,0.41661486698677);
\draw[fill=color0,draw opacity=0] (axis cs:1.6,0) rectangle (axis cs:2.4,0.415646325744935);
\draw[fill=color0,draw opacity=0] (axis cs:2.6,0) rectangle (axis cs:3.4,0.431305657422814);
\draw[fill=color0,draw opacity=0] (axis cs:3.6,0) rectangle (axis cs:4.4,0.440790943049437);
\draw[fill=color0,draw opacity=0] (axis cs:4.6,0) rectangle (axis cs:5.4,0.460917790712668);
\draw[fill=color0,draw opacity=0] (axis cs:5.6,0) rectangle (axis cs:6.4,0.467868712899152);
\draw[fill=color0,draw opacity=0] (axis cs:6.6,0) rectangle (axis cs:7.4,0.480085896269366);
\draw[fill=color0,draw opacity=0] (axis cs:7.6,0) rectangle (axis cs:8.4,0.515851489303949);
\draw[fill=color0,draw opacity=0] (axis cs:8.6,0) rectangle (axis cs:9.4,0.537531091318991);
\draw[fill=color0,draw opacity=0] (axis cs:9.6,0) rectangle (axis cs:10.4,0.592496958541798);
\draw[fill=color0,draw opacity=0] (axis cs:10.6,0) rectangle (axis cs:11.4,0.671824182147869);
\draw[fill=color0,draw opacity=0] (axis cs:11.6,0) rectangle (axis cs:12.4,0.697402865826715);
\draw[fill=color0,draw opacity=0] (axis cs:12.6,0) rectangle (axis cs:13.4,0.770360829396125);
\draw[fill=color0,draw opacity=0] (axis cs:13.6,0) rectangle (axis cs:14.4,0.829706692423229);
\draw[fill=color0,draw opacity=0] (axis cs:14.6,0) rectangle (axis cs:15.4,0.916030567453168);
\draw[fill=color0,draw opacity=0] (axis cs:15.6,0) rectangle (axis cs:16.4,0.977525214406065);
\draw[fill=color0,draw opacity=0] (axis cs:16.6,0) rectangle (axis cs:17.4,1.00148154220955);
\draw[fill=color0,draw opacity=0] (axis cs:17.6,0) rectangle (axis cs:18.4,0.974727360979655);
\draw[fill=color0,draw opacity=0] (axis cs:18.6,0) rectangle (axis cs:19.4,0.847584694279896);
\draw[fill=color0,draw opacity=0] (axis cs:19.6,0) rectangle (axis cs:20.4,0.732774806607501);
\draw[fill=color0,draw opacity=0] (axis cs:20.6,0) rectangle (axis cs:21.4,0.583032669744609);
\draw[fill=color0,draw opacity=0] (axis cs:21.6,0) rectangle (axis cs:22.4,0.449374279715174);
\draw[fill=color0,draw opacity=0] (axis cs:22.6,0) rectangle (axis cs:23.4,0.585848168647253);
\draw[fill=color0,draw opacity=0] (axis cs:23.6,0) rectangle (axis cs:24.4,0.666939038799613);
\draw[fill=color0,draw opacity=0] (axis cs:24.6,0) rectangle (axis cs:25.4,0.874774641140906);
\draw[fill=color0,draw opacity=0] (axis cs:25.6,0) rectangle (axis cs:26.4,1.04323694633995);
\draw[fill=color0,draw opacity=0] (axis cs:26.6,0) rectangle (axis cs:27.4,1.09659098614273);
\draw[fill=color0,draw opacity=0] (axis cs:27.6,0) rectangle (axis cs:28.4,1.07371483415905);
\draw[fill=color3,draw opacity=0] (axis cs:28.6,0) rectangle (axis cs:29.4,1.12841265142064);
\draw[fill=color0,draw opacity=0] (axis cs:29.6,0) rectangle (axis cs:30.4,1.11993041844218);
\draw[fill=color0,draw opacity=0] (axis cs:30.6,0) rectangle (axis cs:31.4,0.987967705091992);
\draw[fill=color0,draw opacity=0] (axis cs:31.6,0) rectangle (axis cs:32.4,1.03789743221177);
\draw[fill=color0,draw opacity=0] (axis cs:32.6,0) rectangle (axis cs:33.4,1.05851743899296);
\draw[fill=color0,draw opacity=0] (axis cs:33.6,0) rectangle (axis cs:34.4,1.07639336582643);
\draw[fill=color0,draw opacity=0] (axis cs:34.6,0) rectangle (axis cs:35.4,1.02686978292214);
\draw[fill=color0,draw opacity=0] (axis cs:35.6,0) rectangle (axis cs:36.4,0.969801641516352);
\draw[fill=color0,draw opacity=0] (axis cs:36.6,0) rectangle (axis cs:37.4,1.01588559560419);
\draw[fill=color0,draw opacity=0] (axis cs:37.6,0) rectangle (axis cs:38.4,0.963116367523077);
\draw[fill=color0,draw opacity=0] (axis cs:38.6,0) rectangle (axis cs:39.4,0.94460697847888);
\draw[fill=color0,draw opacity=0] (axis cs:39.6,0) rectangle (axis cs:40.4,0.911589362286067);
\draw[fill=color0,draw opacity=0] (axis cs:40.6,0) rectangle (axis cs:41.4,0.833482481157829);
\draw[fill=color0,draw opacity=0] (axis cs:41.6,0) rectangle (axis cs:42.4,0.794219635420179);

\end{axis}

\end{tikzpicture}
	\end{minipage}%
	\caption{\textbf{BubbleNets Prediction Sort of Motorbike Video:} The green bar is the annotated training frame selected by BubbleNets. 
	}
	\label{fig:sort_compare}
\end{figure}
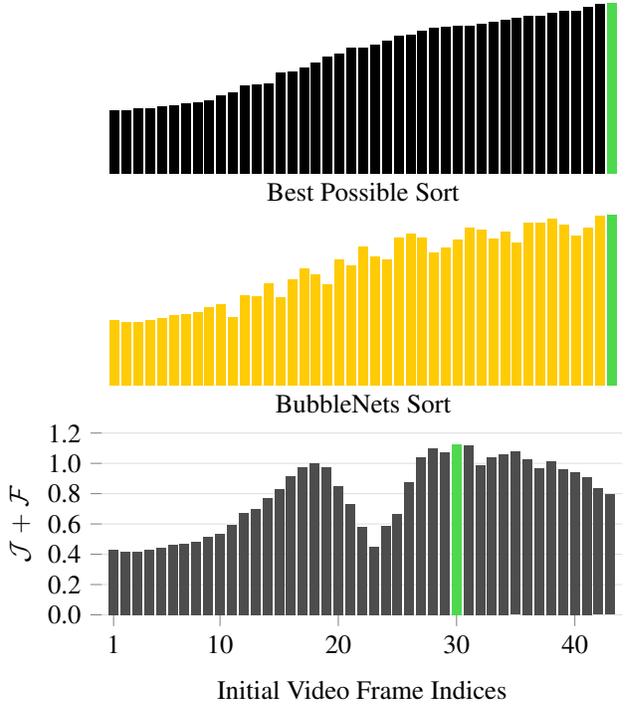

Although BN is not explicitly trained to find the best-performing frame in a video, our complete deep bubble sorting framework is able to accomplish this task, as shown in Figure~\ref{fig:sort_compare}.
Even in cases where the best performing frames are not promoted to the top, an important secondary effect of our deep sorting framework is demoting frames that lead to poorer performance (e.g., Frame 20 in Figure~\ref{fig:frame_performance}); avoiding such frames is critical for annotation frame selection in video object segmentation.

\subsection{BubbleNets Architecture}
\label{sec:BNarch}

Our BubbleNets architecture is shown in Figure~\ref{fig:BN}.
The input has two comparison images, three reference images, and normalized indices \eqref{eq:idx} for all five frames.
Increasing the number of reference frames, $k$, increases video-wide awareness for predicting relative frame performance but also increases network complexity; in practice, we find that $k=$~3 is a good compromise.
The input images are processed using a base Residual Neural Network (ResNet 50, \cite{HeEtAl16}) that is pre-trained on ImageNet, which has been shown to be a good initialization for segmentation~\cite{SFL} and other video tasks \cite{ZhXuCo18}.
Frame indices and ResNet features are fed into BN's performance prediction layers, which consist of four fully-connected layers with decreasing numbers of neurons per layer.
All performance prediction layers include the normalized frame indices as input and use a Leaky ReLU activation function \cite{leakyReLU13}; the later three prediction layers have 20\% dropout for all inputs during training \cite{dropout14}.
After the performance prediction layers, our BN architecture ends with one last fully connected neuron that is the output relative performance prediction $f(x_{i},x_{j},X_{\text{ref.}},\textbf{W})  \in \mathbb{R}$ in \eqref{eq:loss}.

\subsection{BubbleNets Implementation for \\Video Object Segmentation}
\label{sec:BNVOS}

Assume a user wants to segment an object in video and provides an annotation of that object in a single frame.
Because annotating video data is time consuming, we use BubbleNets and deep sorting to automatically select the annotation frame for the user that results in the best segmentation performance possible. 
We segment objects from the annotated frame in the remainder of the video using One-Shot Video Object Segmentation (OSVOS) \cite{OSVOS}.

\subsubsection{Generating Performance Labels for Training}
\label{sec:BNtrain}

Generating performance-based labels to train BN requires a quantitative measure of performance that is measurable on any given video frame.
For our VOS performance measure, we choose a combination of region similarity $\mathcal{J}$ and contour accuracy $\mathcal{F}$.
Region similarity (also known as intersection over union or Jaccard index \cite{jaccard}) provides an intuitive, scale-invariant evaluation for the number of mislabeled foreground pixels with respect to a ground truth annotation.
Given a foreground mask $M$ and ground truth annotation $G$, $\mathcal{J}=\frac{M\cap G}{M \cup G}$.
Contour accuracy evaluates the boundary of a segmentation by measuring differences between the closed set of contours for $M$ and $G$ \cite{DAVIS};
$\mathcal{F}$ is also correlated with  $\mathcal{J}$ \cite[Figure 5]{GrCoWACV2019}.
Using $\mathcal{J}$ and $\mathcal{F}$, we define a frame performance label for loss function \eqref{eq:loss}  as
\begin{align}
	y_i := \frac{1}{n}\sum_{j=1}^{n} \mathcal{J}_j + \mathcal{F}_j,
	\label{eq:label}
\end{align}
where $y_i$ is $i$th label of an $n$-frame video and $\mathcal{J}_j+\mathcal{F}_j$ is the performance on frame $j$ after using frame $i$ for annotation.
In simple words, $y_i$ is the video-wide mean performance that results from selecting the $i$th frame for annotation.

\begin{figure}[t!]
	\centering
	\input{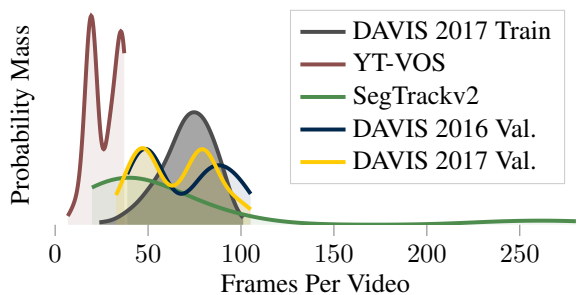}
	\caption{\textbf{PMF for Annotated Frames Available Per Video.}}
	\label{fig:dataset}
\end{figure}

We use our performance label \eqref{eq:label} to generate BN training data.
To avoid labeling costs for annotating BN-selected frames and evaluating segmentation performance, we use a previously annotated VOS dataset.
Our ideal dataset contains many examples and is fully-annotated to provide BN the complete set of video frames for annotation selection.
We give full consideration to the datasets listed in Table~\ref{tab:dataset} \cite{SegTrackv2,DAVIS,DAVIS17,SegTrack,YTVOS}.
YouTube-VOS contains the most annotated frames, but the validation set provides annotations on only the first video frame and the training set provides annotations only on every fifth frame.
SegTrackv2 has the most annotated frames per video, but this metric is skewed by a handful of long videos and the majority of SegTrackv2 videos contain 40 frames or fewer (see Figure~\ref{fig:dataset}).
Accordingly, we use the DAVIS 2017 training set, which contains the most examples of the fully annotated datasets.

Using the DAVIS 2017 training set, we train OSVOS for 500 iterations on every frame and find the resulting performance label \eqref{eq:label}.
For videos with multiple annotated objects, performance labels are generated for each object on every frame.
Preprocessing the dataset takes about a week on a dual-GPU (GTX 1080 Ti) machine but has many benefits.
First, BN can train without running OSVOS, which significantly decreases training time.
Second, we know the ground truth performance of every frame, so we can evaluate the overall deep sorting framework (e.g., seeing which frames are under- or over-promoted in Figure~\ref{fig:sort_compare}).
Finally, we can compare performance against several simple frame selection strategies and know the best and worst frame selections possible for each video in the dataset.

\setlength{\tabcolsep}{2.5pt}
\begin{table}[t]
	\centering
	\caption{
		\textbf{Dataset Metrics:} ``Ant. Frames'' and ``Annotations'' denote the number of annotated frames and object annotations.
		Most SegTrackv2 and all YT-VOS videos have $<$~40 annotated frames.
	}
	\small
	\begin{tabular}{l |c c|c|c|c}
		\hline
		\multicolumn{1}{c|}{} &\multicolumn{2}{c|}{DAVIS 2017} & DAVIS & SegTrack & YT-VOS \\
		\multicolumn{1}{l|}{Number of} &Train & Val. & `16 Val. & v2 & (1st 1,000)\\
		\hline
		\rowcolor{rowgray}	Objects & 144 & 61 & 20 & 24 & 1,000 \\
		Videos & 60 & 30 & 20 & 14 & 607 \\
		\rowcolor{rowgray} Ant. Frames & 4,209 & 1,999 & 1,376 & 1,066 & 16,715  \\
		Annotations & 10,238 & 3,984 & 1,376 & 1,515 & 26,742 \\
		\hline
		\multicolumn{6}{l}{Annotated Frames Per Video} \\
		\hline
		\rowcolor{rowgray}		Mean & 70.2 & 66.6 &  68.8 & 76.1 & 27.5 \\
		Median & 71 & 67.5 & 67.5 & \bf 39 & \bf 30 \\
		\rowcolor{rowgray} \multicolumn{1}{l|}{Range} & 25--100 & 34--104 &  40--104 & 21--279 & 8--36 \\
		\multicolumn{1}{l|}{Coef. of Var.} & 0.22 & 0.31 &  0.32 & 1.03 & 0.29 \\
		\hline
	\end{tabular}
	\label{tab:dataset}
\end{table}

\subsubsection{Five BubbleNets Configurations and Training}
\label{sec:BNconfig}

To test the efficacy of new concepts and establish best practices, we implement five BN configurations for VOS. The first configuration (BN$_0$) uses the standard BN architecture in Section~\ref{sec:BNarch}. The second and third configurations are similar to BN$_0$ but use \textbf{N}o \textbf{I}nput \textbf{F}rame \textbf{I}ndices (BN$_{\text{NIFI}}$) or \textbf{N}o \textbf{R}eference \textbf{F}rames (BN$_{\text{NRF}}$). 
The fourth and fifth configurations are similar to BN$_0$ but use loss functions modified from $\mathcal{L}$ \eqref{eq:loss} that predict \textbf{S}ingle-frame \textbf{P}erformance (BN$_{\mathcal{L}\text{SP}}$) or bias toward middle \textbf{F}rame selection (BN$_{\mathcal{L} \text{F}}$). 

BN$_{\mathcal{L}\text{SP}}$'s single-frame performance loss is defined as:
\begin{align}
\mathcal{L}_{\text{SP}}(\textbf{W}) :=  y_i- f(x_i,X_{\text{ref.}},\textbf{W}),
\label{eq:lossSP}
\end{align}
where $y_i$ is the single performance label for frame $i$.
Alternatively, BN$_{\mathcal{L} \text{F}}$'s middle-frame biased loss is defined as:
\begin{align}
\mathcal{L}_{\text{F}}(\textbf{W}) :=  (y_i-y_j) - ( d_i-d_j ) - f(x_i,x_j,X_{\text{ref.}},\textbf{W}),
\label{eq:lossMF}
\end{align}
where $d_i$ is the distance between frame $i$ and the middle frame. Using the normalized index from \eqref{eq:idx}, we find $d_i$ as:
\begin{align}
d_i = \lambda \left| I_i - I_\text{MF} \right|,
\label{eq:d}
\end{align}
where $I_\text{MF}=0.5$ is the normalized middle frame index and $\lambda=0.5$ determines the relative emphasis of middle frame bias in \eqref{eq:lossMF}.
The intuition behind \eqref{eq:lossMF} is simple.
In addition to predicting the performance difference between frames $i$ and $j$, BN$_{\mathcal{L} \text{F}}$ will learn to consider distance of each frame from the middle of the video.
Given no predicted performance difference, the network will simply fall back on the frame closest to the middle, which is shown in Section~\ref{sec:results} to be an effective annotation choice.
To help BN$_{\mathcal{L} \text{F}}$ learn the additional frame-based loss, we remove all network layer dropout associated with the frame input indices.

\setlength{\tabcolsep}{3pt}
\begin{table}[t!]
	\centering
	\caption{
		\textbf{BubbleNets Configurations.} 
	}
	\small
	\begin{tabular}{l |c c | r| c r | c  c}
		\hline
		\multicolumn{1}{c|}{} & \multicolumn{2}{c|}{Input} & \multicolumn{1}{c|}{} & \multicolumn{2}{c|}{} & \multicolumn{2}{c}{DAVIS `17} \\
		\multicolumn{1}{c|}{Config.} & Frame & Ref. & \multicolumn{1}{c|}{Loss} & \multicolumn{2}{c|}{Total Training} & \multicolumn{2}{c}{Val. Mean} \\
		\multicolumn{1}{c|}{ID} & Index & Frame & \multicolumn{1}{c|}{Funct.} & \multicolumn{2}{l|}{Iterations ~Time} & $\mathcal{J}$ & $\mathcal{F}$ \\
		\hline
		\rowcolor{rowgray}	BN$_0$ & Yes & Yes & $\mathcal{L}$ ~~\eqref{eq:loss} & 3,125 & 5m 11s & \bf 59.7 & \bf 65.5 \\
		BN$_{\text{NIFI}}$ & \bf No & Yes & $\mathcal{L}$ ~~\eqref{eq:loss} & 2,500 & 3m 52s & 58.7 & 65.0 \\
		\rowcolor{rowgray}	BN$_{\mathcal{L}\text{F}}$ & Yes & Yes & \bf $\mathcal{L}_{\text{F}}$ ~\eqref{eq:lossMF} & 8,125 & 15m 30s & 57.8 & 63.8 \\
		BN$_{\text{NRF}}$ & Yes & \bf No & $\mathcal{L}$ ~~\eqref{eq:loss} & 3,125 & 2m 20s & 55.4 & 62.3 \\
		\rowcolor{rowgray} BN$_{\mathcal{L}\text{SP}}$ & Yes & Yes & \bf $\mathcal{L}_{\text{SP}}$ \eqref{eq:lossSP} & 1,875 & 2m 32s & 55.1 & 62.3 \\
		\hline
	\end{tabular}
	\label{tab:config}
\end{table}

All five configurations are trained using the labeled DAVIS 2017 training data described in Section~\ref{sec:BNtrain}.
To decrease training time, all DAVIS 2017 training frames are preprocessed through the ResNet portion of the architecture, which does not change during BN training.
We use a batch size of 1,024 randomly selected videos; each video uses up to five frames that are randomly selected without replacement (e.g., two comparison and three reference).
We add an L1 weight regularization loss with a coefficient of $2\times10^{-6}$, and use the Adam Optimizer \cite{adam14} with a $1\times10^{-3}$ learning rate.
The number of training iterations and training time for each configuration is summarized in Table~\ref{tab:config}.

We evaluate all models using the original bubble sorting framework, although BN$_{\mathcal{L}\text{SP}}$ requires two forward network passes per sort comparison and BN$_{\text{NRF}}$ is deterministic without the random reference frames.
Tasked with learning frame-based loss and frame performance differences, BN$_{\mathcal{L} \text{F}}$ requires the most training iterations of all BN networks.
BN$_{\mathcal{L}\text{SP}}$ trains in fewer iterations due to simplified loss, and both BN$_{\mathcal{L}\text{SP}}$ and BN$_{\text{NRF}}$ train faster due to fewer input images.
As shown in Table~\ref{tab:config}, the BN$_0$ model outperforms BN$_{\mathcal{L}\text{SP}}$ and BN$_{\text{NRF}}$, justifying our claims in Section~\ref{sec:BNrel} for using relative frame performance and reference frames.


\section{Experimental Results}
\label{sec:results}

\subsection{Setup}
Our primary experiments and analysis use the DAVIS 2017 Validation set.
As with the training set in Section~\ref{sec:BNtrain}, we find the segmentation performance for every possible annotated frame, which enables us to do a complete analysis that includes the best and worst possible frame selections and simple frame selection strategies.
We determine the effectiveness of each frame selection strategy by calculating the mean $\mathcal{J}+\mathcal{F}$ for the resulting segmentations on the entire dataset; the mean is calculated on a per video-object basis (e.g., a video with two annotated objects will contribute to the mean twice).
Best and worst frame selections are determined using the combined $\mathcal{J}+\mathcal{F}$ score for each video object.
The simple frame selection strategies are selecting the first frame (current VOS standard), middle frame (found using floor division of video length), last frame, and a random frame from each video for each object.
Finally, because BN results can vary from using random reference frames as input, we only use results from the first run of each configuration (same with random frame selection).

\setlength{\tabcolsep}{10pt}
\begin{table}[t]
	\centering
	\caption{\textbf{Ablation Study on DAVIS 2017 Val. Set:} Study of BN input batch size for bubble sort comparisons and end performance. 
	}
	\small
	\begin{tabular}{c|c c c | c}
		\hline
		\multicolumn{1}{c|}{Batch} & \multicolumn{3}{c|}{Performance ($\mathcal{J}+\mathcal{F}$)} & Mean Video  \\
		\multicolumn{1}{c|}{Size} &  BN$_0$  &  BN$_{\text{NIFI}}$ & BN$_{\mathcal{L}{\text{F}}}$ &  Sort Time \\
		\hline
		\rowcolor{rowgray}		1 & 124.1 & 122.9 & 120.5  & 3.88~s \\ 
		3 & 125.2 & 122.0 & 121.6  & 4.83~s \\ 
		\rowcolor{rowgray}	\bf 5 & \bf 125.2 & \bf 123.8 & \bf 121.7 & 5.32~s \\ 
		10 & 125.2 & 122.0 & 120.3 & 6.52~s \\ 
		\rowcolor{rowgray}		20 & 123.6 & 123.4 & 120.7 & 9.34~s \\ 
		\hline
	\end{tabular}
	\label{tab:ablation}
\end{table}

\subsection{Ablation Study}
\label{sec:ablation}
We perform an ablation study to determine the best batch size for BN predictions.
Recall from Section~\ref{sec:deepsort} that batches reduce variability by using multiple sets of random reference frames.
As shown in Table~\ref{tab:ablation}, a batch size of 5 leads to the best performance for all BN configurations and is chosen as the standard setting for all remaining results.
Also, results varying with batch size for each configuration provides further evidence that reference frames are meaningful for BN's predictions.
Finally, the mean video sort times in Table~\ref{tab:ablation} are for BN$_{\mathcal{L}{\text{F}}}$, which consistently has the highest sort times.
As a practical consideration, we emphasize that the frame selection times in Table~\ref{tab:ablation} are negligible compared to the time it takes a user to annotate a frame \cite{DAVIS2018}.

\setlength{\tabcolsep}{7pt}
\begin{table}[t]
	\centering
	\caption{\textbf{Dataset Annotated Frame Selection Results.} 
	}
	\small
	\begin{tabular}{|l|c|c|c|c|}
		\hline
		\multicolumn{1}{c|}{Annotation} & \multicolumn{4}{c}{Segmentation Performance ($\mathcal{J}+\mathcal{F}$)} \\
		\cline{2-5}
		\multicolumn{1}{c|}{Frame} &  & & & \multicolumn{1}{c}{Coef. of}  \\
		\multicolumn{1}{c|}{Selection} & Mean & Med. & Range & \multicolumn{1}{c}{Variation}   \\
		\hline
		\multicolumn{5}{c}{\bf DAVIS 2017 Val.} \\
		\hline
		\rowcolor{rowbgray} Best  & 141.2 & 143.2 & 14.9--194.9 & 0.26 \\
		\hline
		\bf BN$_0$ & \bf 125.2 & \bf 128.9 & \bf 7.6--194.2 & \bf 0.34 \\
		\rowcolor{rowgray}	\bf BN$_{\text{NIFI}}$ & \bf 123.8 & \bf 129.9 & 8.7--194.2 & \bf 0.35 \\
		\bf BN$_{\mathcal{L}{\text{F}}}$ & \bf 121.7 & \bf 128.0 & \bf 7.6--194.3 & \bf 0.38 \\
		\rowcolor{rowgray}			Middle  & 119.2 & 124.0 & 7.6--193.6 & 0.41  \\
		Random  & 116.5 & 119.7 & 1.6--193.2 & 0.38 \\
		\rowcolor{rowgray}	First  & 113.3 & 117.2 & 3.5--192.5 & 0.39 \\
		Last  & 104.7 & 110.3 & 4.4--190.1 & 0.42 \\
		\hline
		\rowcolor{rowbgray} Worst  & 86.3 & 88.2 & 1.6--188.9 & 0.56 \\
		\hline
		\multicolumn{5}{c}{\bf DAVIS 2016 Val.} \\
		\hline
		\rowcolor{rowbgray}Best  & 171.2 & 176.3 & 130.6--194.9 & 0.11 \\
		\hline
		\bf BN$_0$ & \bf 159.8 & \bf 168.5 & \bf 72.6--194.5 & \bf 0.18 \\
		\rowcolor{rowgray}\bf 	BN$_{\text{NIFI}}$	& \bf 157.3 & \bf 165.7 & \bf 72.6--194.5 & \bf 0.18 \\
		\bf BN$_{\mathcal{L}{\text{F}}}$ & \bf 155.6 & \bf 170.5 & \bf 72.6--193.8 & \bf 0.21 \\
		\rowcolor{rowgray}	Middle  & 155.2 & 169.5 & 77.1--193.8 & 0.21 \\
		First  & 152.8 & 153.4 & 115.2--191.7 & 0.15 \\
		\rowcolor{rowgray}	Random  & 147.5 & 157.3 & 83.1--194.5 & 0.25 \\
		Last  & 147.5 & 153.0 & 72.0--189.6 & 0.23 \\
		\hline
		\rowcolor{rowbgray} Worst  & 127.7 & 141.3 & 68.3--188.9 & 0.31 \\
		\hline
	\end{tabular}
	\label{tab:frameSelect}
\end{table}

\subsection{DAVIS Validation}

Complete annotated frame selection results for the DAVIS 2016 and 2017 validation sets are provided in Table~\ref{tab:frameSelect}.
To put these results in perspective, the current difference in $\mathcal{J}+\mathcal{F}$ for the two leading VOS methods on the DAVIS 2016 Val. benchmark is 2.1 \cite{OSVOS-S,OnAVOS}. 

For first frame selection, it is worth acknowledging that both datasets intend for annotation to take place on the first frame, which guarantees that objects are visible for annotation (in some videos, objects become occluded or leave the view).
Despite this advantage, middle frame selection outperforms first frame selection on both datasets overall and on 3/5 of the videos on DAVIS 2017 Val.
In fact, on both datasets first frame selection is, on average, closer to the worst possible frame selection than the best.
Last frame selection has the worst performance and, using the coefficient of variation, the most variable relative performance.
Finally, the best performing annotation frame is never the first or last frame for any DAVIS validation video.

Middle frame selection has the best performance of all simple strategies.
We believe that the intuition for this is simple.
Because the middle frame has the least cumulative temporal distance from all other frames, it is on average more representative of the other frames with respect to annotated object positions and poses.
Thus, the middle frame is, on average, the best performing frame for segmentation.


\begin{figure}[t!]
	\centering
	\input{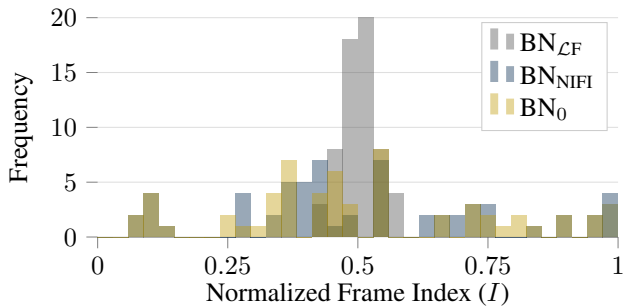}
	\caption{\textbf{Frame-Selection Locations in Video:} Normalized indices \eqref{eq:idx} of all BN annotation frame selections on DAVIS `17 Val.}
	\label{fig:idxselect}
\end{figure}

All BN configurations outperform the simple selection strategies, and BN$_0$ performs best of all BN configurations.
When selecting different frames, BN$_0$ beats middle frame selection on 3/5 videos  and first frame selection on 4/5 videos for DAVIS 2017 Val.
By comparing the performance of BN$_0$ and BN$_{\text{NIFI}}$, we find that BN$_0$'s use of normalized frame indices \eqref{eq:idx} is beneficial for performance.

Finally, it is clear from the frame-selection locations in Figure~\ref{fig:idxselect} that BN$_{\mathcal{L}{\text{F}}}$'s  modified loss function \eqref{eq:lossMF} successfully biases selections toward the middle of each video. 

\setlength{\tabcolsep}{7pt}
\begin{table}[t!]
	\centering
	\caption{\textbf{Results on Datasets with Limited Frames Per Video.} 
	}
	\small
	\begin{tabular}{|l|c|c|c|c|}
		\hline
		\multicolumn{1}{c|}{Annotation} & \multicolumn{4}{c}{Segmentation Performance ($\mathcal{J}+\mathcal{F}$)} \\
		\cline{2-5}
		\multicolumn{1}{c|}{Frame} &  & & & \multicolumn{1}{c}{Coef. of}  \\
		\multicolumn{1}{c|}{Selection} & Mean & Med. & Range & \multicolumn{1}{c}{Variation}   \\
		\hline
		\multicolumn{5}{c}{\bf SegTrackv2} \\
		\hline
		\bf BN$_{\mathcal{L}{\text{F}}}$ & \bf 134.7 & \bf 145.9 & \bf 14.3--184.6 & \bf 0.32 \\
		\rowcolor{rowgray}	Middle  & 134.5 & 143.5 & 14.3--182.8 & 0.32 \\
		BN$_{\text{NIFI}}$	& 134.3 & 144.2 & 33.9--178.5 & 0.30 \\
		\rowcolor{rowgray}BN$_0$ & 130.6 & 127.3 & 50.0--183.2 & 0.30 \\
		Last  & 123.6 & 130.4 & 14.3--178.4 & 0.36 \\
		\rowcolor{rowgray} First  & 122.3 & 122.5 & 45.8--181.7 & 0.31 \\
		\hline
		\multicolumn{5}{c}{ \textbf{YT-VOS} (1st 1,000)} \\
		\hline
		\bf BN$_{\mathcal{L}{\text{F}}}$ & \bf115.5 & \bf126.6 & \bf0.0--197.3 & \bf0.46 \\
		\rowcolor{rowfgray}	Middle & 115.0 & 124.2 & 0.0--196.2 & 0.46 \\
		BN$_{\text{NIFI}}$ & 111.8 & 121.0 & 0.0--196.3 & 0.47 \\
		\rowcolor{rowfgray}BN$_0$ & 110.4 & 121.5 & 0.0--194.1 & 0.49 \\
		First  & 107.3 & 114.0 & 0.0--196.3 & 0.49 \\
		\rowcolor{rowfgray}	 Last & 101.2 & 108.1 & 0.0--195.4 & 0.56 \\
		\hline
	\end{tabular}
	\label{tab:frameSelectLimited}
\end{table}

\subsection{Results on Datasets with Limited Frames}

Annotated frame selection results for SegTrackv2 and YouTube-VOS are provided in Table~\ref{tab:frameSelectLimited}.
As emphasized in Section~\ref{sec:BNtrain}, the videos in these datasets have a limited number of frames available for annotation, which limits the effectiveness of BN frame selection.
Because the YouTube-VOS validation set only provides annotations on the first frame, we instead evaluate on the first 1,000 objects of the YouTube-VOS training set, which provides annotations on every fifth frame.
This reduces the number of candidate annotation frames that BN can compare, sort, and select to one fifth of that available in a standard application for the same videos.
While all BN configurations outperform first and last frame selection, BN$_{\mathcal{L}{\text{F}}}$ is the only configuration that consistently outperforms all other selection strategies.
We postulate that the additional bias of BN$_{\mathcal{L}{\text{F}}}$ toward index-based selections made this configuration more robust to reductions in candidate annotation frames.

\begin{table}
	\centering
	\caption{\textbf{Cross Evaluation of Benchmark Methods:} OSVOS and OnAVOS DAVIS `17 Val. results using identical frame selections.}
	\small
	\begin{tabular}{l | c | c | c | c | c}
		\hline
		\multicolumn{1}{c|}{Segmentation} & \multicolumn{5}{c}{Frame Selection and DAVIS $\mathcal{J}$ \& $\mathcal{F}$ Mean} \\
		\cline{2-6}
		\multicolumn{1}{c|}{Method} & First & Middle & BN$_{\mathcal{L}\text{F}}$ & BN$_{\text{NIFI}}$ & BN$_0$  \\
		\hline
		\rowcolor{rowgray} OSVOS & 56.6 & 59.6 & 60.8 & 61.9 & \bf 62.6 \\
		OnAVOS & 63.9 & 68.4 & 68.5 & 68.4 & \bf 69.2 \\
		\hline
	\end{tabular}
	\label{tab:crosseval}
\end{table}

\subsection{Results on Different Segmentation Methods}

Cross evaluation results for different segmentation methods are provided in Table~\ref{tab:crosseval}.
All BN configurations select annotation frames that improve the performance of OnAVOS, despite BN training exclusively on OSVOS-generated labels. 
Nonetheless, the label-generating formulation in Section~\ref{sec:BNtrain} is general to other semi-supervised VOS methods; thus, new BN training labels can always be generated for other methods.
Note that first frame results in Table~\ref{tab:crosseval} differ from the online benchmark due to dataset-specific configurations (e.g., \cite{OnAVOS17}), non-deterministic components, and our segmenting and evaluating objects from multi-object videos separately (i.e., we do not  combine multi-object video masks for a single, non-overlapping evaluation). 

\subsection{Final Considerations for Implementation}


\begin{figure}[t!]
	\centering
	\includegraphics[width=0.475\textwidth]{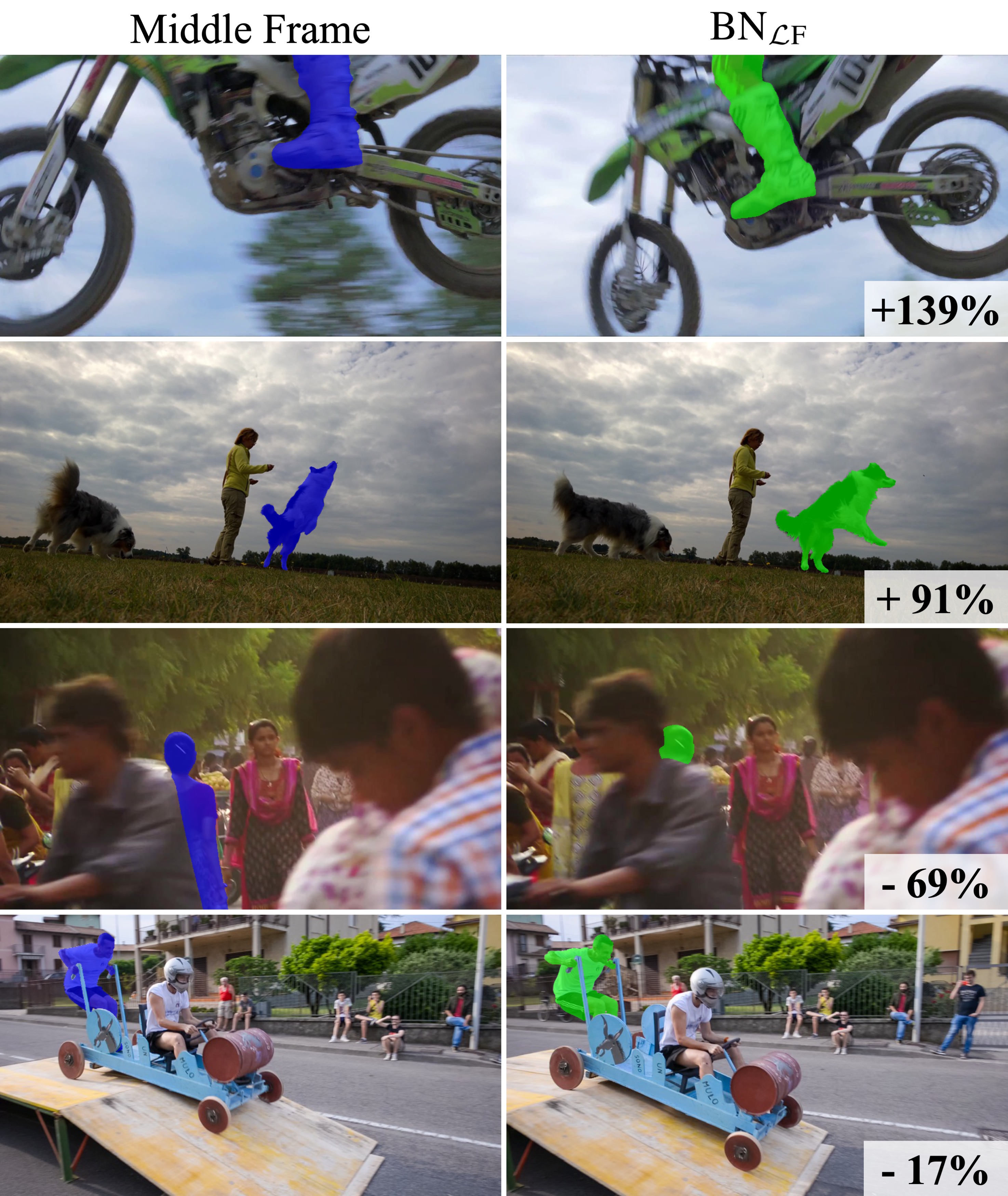}
	\caption{\textbf{BN$_{\mathcal{L}\text{F}}$--Middle Frame Comparison:} Two best (top) and worst (bottom) BN$_{\mathcal{L}\text{F}}$ selections relative to the middle frame.
	}
	\label{fig:BNLF}
\end{figure}

Selecting the middle frame for annotation is the best performing simple selection strategy on all datasets and is easy to implement in practice.
However, BN$_{\mathcal{L}\text{F}}$ is more reliable than middle frame selection and results in better segmentation performance on all datasets.
As shown in Figure~\ref{fig:idxselect}, BN$_{\mathcal{L}\text{F}}$ selects frames close to the middle of each video, but deviates toward frames that, on average, result in better performance than the middle frame (see Tables~\ref{tab:frameSelect}~\&~\ref{tab:frameSelectLimited}).
On DAVIS 2017 Val., BN$_{\mathcal{L}\text{F}}$ deviations from the middle frame results in better performance 70\% of the time.
We believe the underlying mechanism for this improvement is recognizing when the middle frame exhibits less distinguishable ResNet features or is less representative of the video reference frames.
To demonstrate beneficial and counterproductive examples of this behavior, the two best and worst BN$_{\mathcal{L}\text{F}}$ selections relative to the middle frame on DAVIS 2017 Val. are shown with relative performance \%'s in Figure~\ref{fig:BNLF}.

\begin{figure*}[t!]
	\centering
	\includegraphics[width=0.975\textwidth]{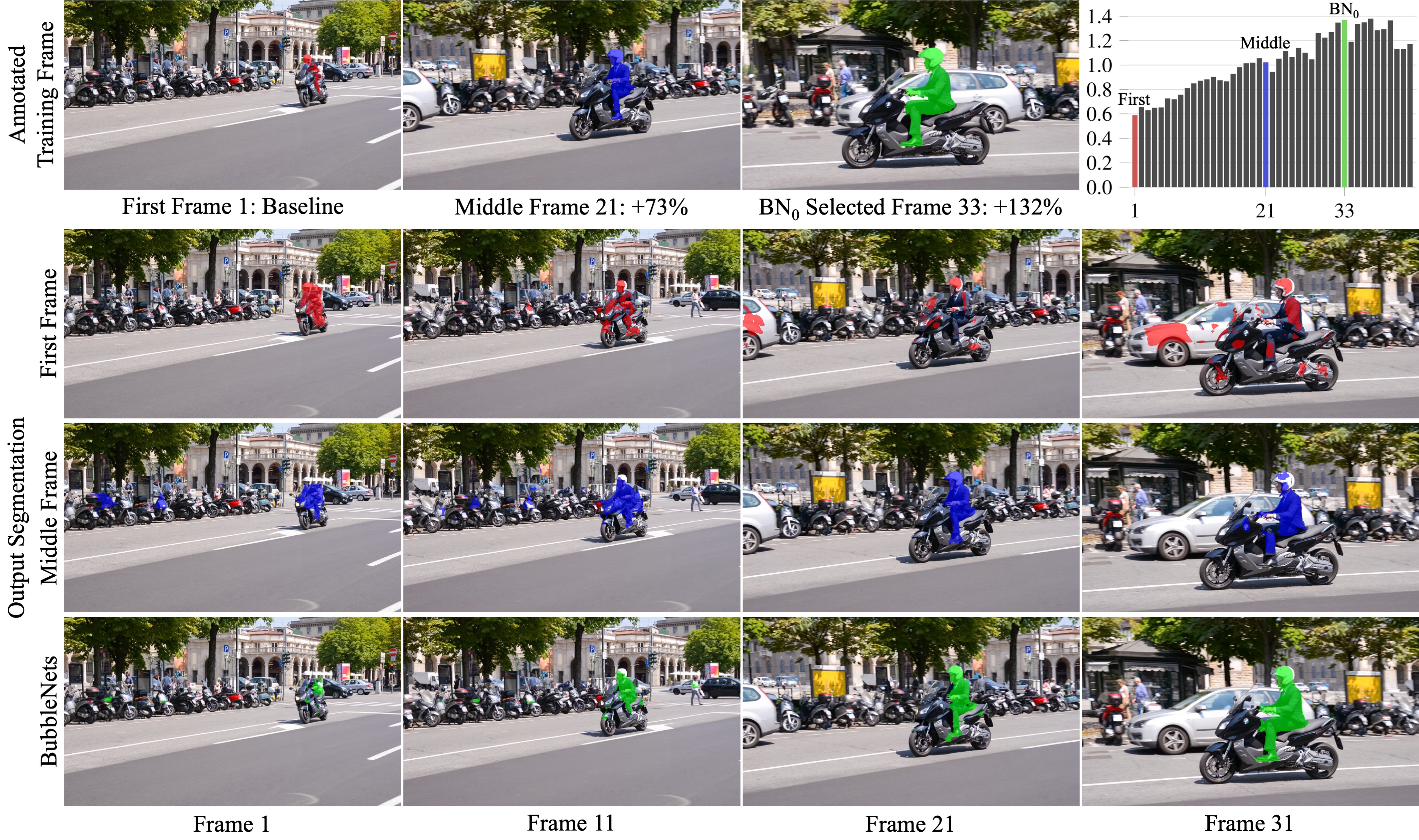}
	\caption{\textbf{Qualitative Comparison on DAVIS 2017 Validation Set:} Segmentations from different annotated frame selection strategies.
	}
	\label{fig:BN_performance}
\end{figure*}

BN$_0$ has the greatest relative segmentation improvements over simple selection strategies on the DAVIS validation datasets (see example comparison in Figure~\ref{fig:BN_performance}).
However, this performance did not translate to datasets with a limited number of annotation frames available.
To determine if this is due to domain shift of fewer frames, we analyze the 10 longest and shortest videos from DAVIS 2017 Val. in Table~\ref{tab:long_short} as an additional experiment.
The key result is that BN$_0$ and BN$_{\text{NIFI}}$'s relative performance gains double once approximately forty annotation frames are available.
This is encouraging as most real-world videos have many more frames available for annotation, which is conducive for BN$_0$'s best annotated frame selection results.

\setlength{\tabcolsep}{5pt}
\begin{table}[t!]
	\centering
	\caption{\textbf{Frames Per Video and Relative Performance:} BN performances relative to first frame on DAVIS 2017 Validation.
	}
	\small
	\begin{tabular}{l| c | l l l }
		\hline
		\multicolumn{1}{c|}{Videos from} & Number of & \multicolumn{3}{c}{Relative Mean ($\mathcal{J}+\mathcal{F}$)}   \\
		\multicolumn{1}{c|}{DAVIS `17 Val.} & Frames &  \multicolumn{1}{c}{BN$_0$}  &  \multicolumn{1}{c}{BN$_{\text{NIFI}}$} & \multicolumn{1}{c}{BN$_{\mathcal{L}{\text{F}}}$} \\
		\hline
		\rowcolor{rowgray} 10 Longest & 81--104 & \bf +~11.8\% & \bf +~10.9\% & +~4.0\% \\ 
		All & 34--104 & \bf +~10.5\% & \bf +~~~9.3\% & +~7.4\% \\ 
		\rowcolor{rowgray} 10 Shortest & \bf 34--43~~ & +~~~4.9\% & +~~~5.0\% & +~3.3\% \\ 
		\hline
	\end{tabular}
	\label{tab:long_short}
\end{table}

\section{Conclusions}

We emphasize that automatic selection of the best-performing annotation frames for video object segmentation is a \emph{hard} problem.
Still, as video object segmentation methods become more learning-based and data-driven, it is critical that we make the most of training data and users' time for annotation.
The most recent DAVIS challenge has shifted focus toward improving performance given limited annotation feedback \cite{DAVIS2018}.
However, we demonstrate in this work that there are already simple strategies available that offer a significant performance improvement over first frame annotations without increasing user effort; likewise, our BubbleNets framework further improves performance using learned annotated frame selection.
To continue progress in this direction and improve video object segmentation algorithms in practice, dataset annotators should give full consideration to alternative frame selection strategies when preparing future challenges.

Finally, while the current BubbleNets implementation is specific to video object segmentation, it is more widely applicable.
In future work, we plan to apply BubbleNets to improve performance in other video-based applications.

\section*{Acknowledgements}
\noindent This work was partially supported by the DARPA MediFor program under contract FA8750-16-C-0168.

{\small
\bibliographystyle{ieee}
\bibliography{GrCoCVPR2019_BN}

\begin{thebibliography}{10}\itemsep=-1pt

\bibitem{SEA}
S.~Avinash~Ramakanth and R.~Venkatesh~Babu.
\newblock Seamseg: Video object segmentation using patch seams.
\newblock In {\em Proceedings of the IEEE Conference on Computer Vision and
  Pattern Recognition (CVPR)}, 2014.

\bibitem{CINM}
L.~Bao, B.~Wu, and W.~Liu.
\newblock {CNN} in {MRF:} video object segmentation via inference in {A}
  cnn-based higher-order spatio-temporal {MRF}.
\newblock In {\em IEEE Conference on Computer Vision and Pattern Recognition
  (CVPR)}, 2018.

\bibitem{BeEtAl18}
J.~Bernard, M.~Hutter, M.~Zeppelzauer, D.~Fellner, and M.~Sedlmair.
\newblock Comparing visual-interactive labeling with active learning: An
  experimental study.
\newblock {\em IEEE Transactions on Visualization and Computer Graphics},
  24(1):298--308, Jan 2018.

\bibitem{OSVOS}
S.~Caelles, K.-K. Maninis, J.~Pont-Tuset, L.~Leal-Taix\'e, D.~Cremers, and
  L.~{Van Gool}.
\newblock One-shot video object segmentation.
\newblock In {\em IEEE Conference on Computer Vision and Pattern Recognition
  (CVPR)}, 2017.

\bibitem{DAVIS2018}
S.~Caelles, A.~Montes, K.~Maninis, Y.~Chen, L.~V. Gool, F.~Perazzi, and
  J.~Pont{-}Tuset.
\newblock The 2018 {DAVIS} challenge on video object segmentation.
\newblock {\em CoRR}, abs/1803.00557, 2018.

\bibitem{CaZi17}
J.~Carreira and A.~Zisserman.
\newblock Quo vadis, action recognition? a new model and the kinetics dataset.
\newblock In {\em 2017 IEEE Conference on Computer Vision and Pattern
  Recognition (CVPR)}, 2017.

\bibitem{PML}
Y.~Chen, J.~Pont-Tuset, A.~Montes, and L.~{Van Gool}.
\newblock Blazingly fast video object segmentation with pixel-wise metric
  learning.
\newblock In {\em Computer Vision and Pattern Recognition (CVPR)}, 2018.

\bibitem{FAVOS}
J.~Cheng, Y.-H. Tsai, W.-C. Hung, S.~Wang, and M.-H. Yang.
\newblock Fast and accurate online video object segmentation via tracking
  parts.
\newblock In {\em IEEE Conference on Computer Vision and Pattern Recognition
  (CVPR)}, 2018.

\bibitem{SFL}
J.~Cheng, Y.-H. Tsai, S.~Wang, and M.-H. Yang.
\newblock Segflow: Joint learning for video object segmentation and optical
  flow.
\newblock In {\em IEEE International Conference on Computer Vision (ICCV)},
  2017.

\bibitem{ImageNet}
J.~Deng, W.~Dong, R.~Socher, L.~J. Li, K.~Li, and L.~Fei-Fei.
\newblock Imagenet: A large-scale hierarchical image database.
\newblock In {\em IEEE Conference on Computer Vision and Pattern Recognition
  (CVPR)}, 2009.

\bibitem{jaccard}
M.~Everingham, L.~Van~Gool, C.~K. Williams, J.~Winn, and A.~Zisserman.
\newblock The pascal visual object classes ({VOC}) challenge.
\newblock {\em International journal of computer vision}, 88(2):303--338, 2010.

\bibitem{NLC}
A.~Faktor and M.~Irani.
\newblock Video segmentation by non-local consensus voting.
\newblock In {\em British Machine Vision Conference (BMVC)}, 2014.

\bibitem{JMP}
Q.~Fan, F.~Zhong, D.~Lischinski, D.~Cohen-Or, and B.~Chen.
\newblock Jumpcut: non-successive mask transfer and interpolation for video
  cutout.
\newblock {\em ACM Trans. Graph.}, 34(6):195, 2015.

\bibitem{GrCoWACV2019}
B.~A. Griffin and J.~J. Corso.
\newblock Tukey-inspired video object segmentation.
\newblock In {\em IEEE Winter Conference on Applications of Computer Vision
  (WACV)}, 2019.

\bibitem{HeEtAl16}
K.~He, X.~Zhang, S.~Ren, and J.~Sun.
\newblock Deep residual learning for image recognition.
\newblock In {\em The IEEE Conference on Computer Vision and Pattern
  Recognition (CVPR)}, 2016.

\bibitem{VPN}
V.~Jampani, R.~Gadde, and P.~V. Gehler.
\newblock Video propagation networks.
\newblock In {\em IEEE Conference on Computer Vision and Pattern Recognition
  (CVPR)}, 2017.

\bibitem{CTN}
W.~D. Jang and C.~S. Kim.
\newblock Online video object segmentation via convolutional trident network.
\newblock In {\em 2017 IEEE Conference on Computer Vision and Pattern
  Recognition (CVPR)}, pages 7474--7483, July 2017.

\bibitem{DAVIS2017-2nd}
A.~Khoreva, R.~Benenson, E.~Ilg, T.~Brox, and B.~Schiele.
\newblock Lucid data dreaming for object tracking.
\newblock {\em The 2017 DAVIS Challenge on Video Object Segmentation - CVPR
  Workshops}, 2017.

\bibitem{adam14}
D.~P. Kingma and J.~Ba.
\newblock Adam: {A} method for stochastic optimization.
\newblock In {\em International Conference on Learning Representations (ICLR)},
  2014.

\bibitem{KN98}
D.~Knuth.
\newblock {\em {The Art of Computer Programming}}, volume 1-3.
\newblock Addison-Wesley Longman Publishing Co., Inc., Boston, MA, USA, 1998.

\bibitem{KEY}
Y.~J. Lee, J.~Kim, and K.~Grauman.
\newblock Key-segments for video object segmentation.
\newblock In {\em IEEE International Conference on Computer Vision (ICCV)},
  2011.

\bibitem{SegTrackv2}
F.~Li, T.~Kim, A.~Humayun, D.~Tsai, and J.~M. Rehg.
\newblock Video segmentation by tracking many figure-ground segments.
\newblock In {\em The IEEE International Conference on Computer Vision (ICCV)}.

\bibitem{DAVIS2017-1st}
X.~Li, Y.~Qi, Z.~Wang, K.~Chen, Z.~Liu, J.~Shi, P.~Luo, C.~C. Loy, and X.~Tang.
\newblock Video object segmentation with re-identification.
\newblock {\em The 2017 DAVIS Challenge on Video Object Segmentation - CVPR
  Workshops}, 2017.

\bibitem{leakyReLU13}
A.~L. Maas, A.~Y. Hannun, and A.~Y. Ng.
\newblock Rectifier nonlinearities improve neural network acoustic models.
\newblock In {\em ICML Workshop on Deep Learning for Audio, Speech and Language
  Processing}, 2013.

\bibitem{OSVOS-S}
K.~Maninis, S.~Caelles, Y.~Chen, J.~Pont-Tuset, L.~Leal-Taixé, D.~Cremers, and
  L.~V. Gool.
\newblock Video object segmentation without temporal information.
\newblock {\em IEEE Transactions on Pattern Analysis and Machine Intelligence},
  pages 1--1, 2018.

\bibitem{ColdStart98}
A.~McCallum and K.~Nigam.
\newblock Employing {EM} and pool-based active learning for text
  classification.
\newblock In {\em In International Conference on Machine Learning (ICML)},
  1998.

\bibitem{BVS}
N.~Märki, F.~Perazzi, O.~Wang, and A.~Sorkine-Hornung.
\newblock Bilateral space video segmentation.
\newblock In {\em 2016 IEEE Conference on Computer Vision and Pattern
  Recognition (CVPR)}, pages 743--751, June 2016.

\bibitem{RGMP}
S.~W. Oh, J.-Y. Lee, K.~Sunkavalli, and S.~J. Kim.
\newblock Fast video object segmentation by reference-guided mask propagation.
\newblock In {\em IEEE Conference on Computer Vision and Pattern Recognition
  (CVPR)}, 2018.

\bibitem{FST}
A.~Papazoglou and V.~Ferrari.
\newblock Fast object segmentation in unconstrained video.
\newblock In {\em Proceedings of the IEEE International Conference on Computer
  Vision (ICCV)}, 2013.

\bibitem{MSK}
F.~Perazzi, A.~Khoreva, R.~Benenson, B.~Schiele, and A.Sorkine-Hornung.
\newblock Learning video object segmentation from static images.
\newblock In {\em IEEE Conference on Computer Vision and Pattern Recognition
  (CVPR)}, 2017.

\bibitem{DAVIS}
F.~Perazzi, J.~Pont-Tuset, B.~McWilliams, L.~Van~Gool, M.~Gross, and
  A.~Sorkine-Hornung.
\newblock A benchmark dataset and evaluation methodology for video object
  segmentation.
\newblock In {\em IEEE Conference on Computer Vision and Pattern Recognition
  (CVPR)}, 2016.

\bibitem{FCP}
F.~Perazzi, O.~Wang, M.~Gross, and A.~Sorkine-Hornung.
\newblock Fully connected object proposals for video segmentation.
\newblock In {\em IEEE International Conference on Computer Vision (ICCV)},
  2015.

\bibitem{DAVIS17}
J.~Pont-Tuset, F.~Perazzi, S.~Caelles, P.~Arbel\'aez, A.~Sorkine-Hornung, and
  L.~{Van Gool}.
\newblock The 2017 davis challenge on video object segmentation.
\newblock {\em arXiv:1704.00675}, 2017.

\bibitem{Bu12}
B.~Settles.
\newblock Active learning.
\newblock {\em Synthesis Lectures on Artificial Intelligence and Machine
  Learning}, 6(1):1--114, 2012.

\bibitem{PLM}
J.~Shin~Yoon, F.~Rameau, J.~Kim, S.~Lee, S.~Shin, and I.~So~Kweon.
\newblock Pixel-level matching for video object segmentation using
  convolutional neural networks.
\newblock In {\em IEEE International Conference on Computer Vision (ICCV)},
  2017.

\bibitem{dropout14}
N.~Srivastava, G.~Hinton, A.~Krizhevsky, I.~Sutskever, and R.~Salakhutdinov.
\newblock Dropout: A simple way to prevent neural networks from overfitting.
\newblock {\em Journal of Machine Learning Research}, 15:1929--1958, 2014.

\bibitem{SegTrack}
D.~Tsai, M.~Flagg, A.~Nakazawa, and J.~M. Rehg.
\newblock Motion coherent tracking using multi-label mrf optimization.
\newblock {\em International journal of computer vision}, 100(2):190--202,
  2012.

\bibitem{ViGr09}
S.~Vijayanarasimhan and K.~Grauman.
\newblock What's it going to cost you?: Predicting effort vs. informativeness
  for multi-label image annotations.
\newblock In {\em IEEE Conference on Computer Vision and Pattern Recognition
  (CVPR)}, 2009.

\bibitem{ViGr14}
S.~Vijayanarasimhan and K.~Grauman.
\newblock Large-scale live active learning: Training object detectors with
  crawled data and crowds.
\newblock {\em International Journal of Computer Vision}, 108(1):97--114, May
  2014.

\bibitem{ViJaGr10}
S.~Vijayanarasimhan, P.~Jain, and K.~Grauman.
\newblock Far-sighted active learning on a budget for image and video
  recognition.
\newblock In {\em IEEE Conference on Computer Vision and Pattern Recognition
  (CVPR)}, 2010.

\bibitem{OnAVOS17}
P.~Voigtlaender and B.~Leibe.
\newblock Online adaptation of convolutional neural networks for the 2017 davis
  challenge on video object segmentation.
\newblock {\em The 2017 DAVIS Challenge on Video Object Segmentation - CVPR
  Workshops}, 2017.

\bibitem{OnAVOS}
P.~Voigtlaender and B.~Leibe.
\newblock Online adaptation of convolutional neural networks for video object
  segmentation.
\newblock In {\em British Machine Vision Conference (BMVC)}, 2017.

\bibitem{VoRa11}
C.~Vondrick and D.~Ramanan.
\newblock Video annotation and tracking with active learning.
\newblock In J.~Shawe-Taylor, R.~S. Zemel, P.~L. Bartlett, F.~Pereira, and
  K.~Q. Weinberger, editors, {\em Advances in Neural Information Processing
  Systems 24}, pages 28--36. Curran Associates, Inc., 2011.

\bibitem{WeSz17}
S.~Wehrwein and R.~Szeliski.
\newblock Video segmentation with background motion models.
\newblock In {\em British Machine Vision Conference (BMVC)}, 2017.

\bibitem{YTVOS}
N.~Xu, L.~Yang, Y.~Fan, D.~Yue, Y.~Liang, J.~Yang, and T.~Huang.
\newblock Youtube-vos: A large-scale video object segmentation benchmark.
\newblock {\em arXiv preprint arXiv:1809.03327}, 2018.

\bibitem{OSMN}
L.~Yang, Y.~Wang, X.~Xiong, J.~Yang, and A.~K. Katsaggelos.
\newblock Efficient video object segmentation via network modulation.
\newblock {\em IEEE Conference on Computer Vision and Pattern Recognition
  (CVPR)}, 2018.

\bibitem{ZhXuCo18}
L.~Zhou, C.~Xu, and J.~Corso.
\newblock Towards automatic learning of procedures from web instructional
  videos.
\newblock In {\em AAAI Conference on Artificial Intelligence}, 2018.

\end{thebibliography}
}

\end{document}